\documentclass[conference]{IEEEtran}
\usepackage[table, dvipsnames]{xcolor}
\usepackage{lipsum} 				\usepackage[margin = 1in, left =1.5in,includefoot]{geometry}
\usepackage[hidelinks]{hyperref}		\usepackage{enumerate}

\usepackage{hyperref}
\hypersetup{
    colorlinks=true,
    linkcolor=blue,
    filecolor=magenta,      
    urlcolor=cyan,
}

\usepackage{listings}
\usepackage{graphicx}		
\usepackage{float}			
\usepackage{mhchem}		
\usepackage{fancyhdr}
\usepackage{algorithmicx}
\usepackage{algpseudocode}
\algdef{SE}[DOWHILE]{Do}{doWhile}{\algorithmicdo}[1]{\algorithmicwhile\ #1}%

\usepackage[english]{babel}
\usepackage[utf8]{inputenc}
\usepackage{amsmath}
\usepackage{amsfonts}
\usepackage{graphicx}
\usepackage[colorinlistoftodos]{todonotes}
\usepackage{algorithm}
\usepackage{algpseudocode}

\usepackage{geometry}
\ifCLASSINFOpdf
\else
\fi
\begin{document}

\begin{titlepage}

\begin{center}
\baselineskip=14pt

\vspace*{5.7cm}
{\large\bf Clustering and Labelling Auction Fraud Data}\\
\baselineskip=14pt
\vspace*{6pt}

{\large Ahmad Alzahrani and Samira Sadaoui}\\ % use long format for names as in Cristina E. Manfredotti and put a slash between the initials (if any) and the surname
\vspace*{6pt}
{\large CS 2018-08}\\ % e.g., the sixth technical report of the year 2025
{\large DOI: 10.6084/m9.figshare.6993308}
\vspace*{4.5cm}

Copyright \copyright{} 2018, A.\ Alzahrani, S.\ Sadaoui\\ % use the short format for names, e.g., initials of given names followed by surname and put a slash between the initials and the surname as in C.E.\ Manfredotti and H.J.\ Hamilton
Department of Computer Science\\
University of Regina\\
Regina, SK, CANADA\\
S4S 0A2\\
%\vspace{1cm}
%{\large
%ISSN 0828-3494\\ % this values stays the same for every tech report
%ISBN xyz-0-xyzq-xyzq-t (print)\\
%ISBN xyz-0-xyzq-xyzq-t (on-line)}\\
\end{center}
%\vfill 
\end{titlepage}

\begin{abstract}
Although shill bidding is a common auction fraud, it is however very tough to detect. Due to the unavailability and lack of training data, in this study, we build a high-quality labeled shill bidding dataset based on recently collected auctions from eBay.  Labeling shill biding instances with multidimensional features is a critical phase for the fraud classification task.  For this purpose, we introduce a new approach to systematically label the fraud data with the help of the hierarchical clustering CURE that returns remarkable results as illustrated in the experiments. 
\end{abstract}

\IEEEpeerreviewmaketitle

\section{Introduction}
The last three decades witnessed a significant increase in exchanging goods and services over the Web.  According to the World Trade Organization, the worldwide merchandise during the period 1995-2015 was over 18 billion \footnote{https://www.wto.org/english/res\_e/statis\_e/its2015\_e/its2015\_e.pdf}. Online auctions are a very profitable e-commerce application. In 2017, eBay claimed that the net revenue reached 9.7 billion US dollars, and the number of active users hit 170 million \footnote{https://www.statista.com}. Regardless of their popularity, e-auctions remain very vulnerable to cyber-crimes. The high anonymity of users, low fees of auction services and flexibility of bidding make auctions a great incubator for fraudulent activities. The Internet Crime Complain Center (IC3) announced that auction fraud is one of the top cyber-crimes \cite{CI32016}.  As an example, the complaints about auction fraud in only three states, California, Florida and New York, reached 7,448 in 2016 \cite{CI32016}. Fraudsters can commit three types of fraud, which are pre-auction fraud, such as auctioning of black market merchandise, in-auction fraud that occurs during the bidding time, such as Shill Bidding (SB), and post-auction fraud, such as fees stacking.  Our primary focus is on SB whose goal is to increase the profits of sellers. SB does not leave any concrete evidence unlike the two other fraud. Indeed, buyers are not even aware that they have been overcharged.

\begin{figure*}[h]
\centering
\includegraphics[scale=.6]{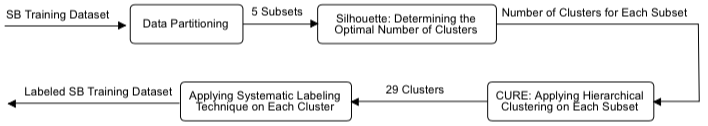} 
\caption{The Labeling Process of the SB Training Dataset}
\label{Labeling_Diagram}
\end{figure*}

Identifying relevant SB strategies, determining robust SB metrics, preprocessing commercial auction data, and finally evaluation  the SB metrics based on the extracted data make the study of SB fraud very challenging.  In addition, labeling SB instances with multidimensional features is a critical phase for the classification models. In the literature, labeling training data is usually done manually by the domain experts, which is quite a laborious task and is prone to errors too. Due to the unavailability and lack of labeled SB training datasets, the main contribution of this paper is to produce high-quality labeled SB data based on commercial auction transactions that we have extracted lately and preprocessed \cite{alzahrani2018scraping}. As illustrated  in Figure \ref{Labeling_Diagram}, we introduce a new approach to systematically label SB data with the help of data clustering. This approach consists of splitting first the SB dataset into several subsets according to the different bidding durations.

Since hierarchical clustering is significantly preferable over partitioning clustering and provides a higher quality of clusters, we applied the Clustering Using Representatives (CURE) technique to produce the best differentiation between normal and suspicious bidders.  CURE \cite{guha1998cure} has been proved over the years to be a very efficient clustering method for large-scale training datasets in terms of the cluster quality and outlier elimination. Hierarchical clustering has been practiced successfully in numerous fraud studies \cite{ford2012A, ganguly2018online}.    The resulting labeled SB dataset can be employed by the state-of-the-art classification methods. Furthermore, the accuracy of new predictive models can also be tested using our SB training dataset. 

\section{Shill Bidding Overview}
SB is a well-known auction fraud, and yet it is the most difficult to detect since it behaves similarly to normal bidding \cite{dong2009combating, ford2012A}. It aim is to increase the price or desirability of the auctioned item. In other words, the objective of a shill bidder is not to win the auction but to raise the revenue of the seller. SB leads buyers to overpay for the items, especially for high priced items. Thus, e-auctions may lose their credibility \cite{dong2009combating}. In Table \ref{SBPatterns}, we highlight the most relevant SB patterns \cite{Sadaoui2017}.  Each pattern, representing a classification feature,  provides a unique aspect of the bidding behaviour, which is very important for building robust fraud classifiers.

\begin{table}[h]
	\centering
	\caption{SB Patterns and their Characteristics}
	\begin{tabular}{|p{0.5in}|p{1.3in}|p{.4in}|p{.4in}|}
		\hline
			{\bf Name} & {\bf Definition} & {\bf Category} & {\bf Source} \\ \hline
			Bidder Tendency (BT)& Participates exclusively in auctions of few sellers rather than a diversified lot & Bidder & User history\\ \hline
			Bidding Ratio (BR)& Participates more frequently to raise the auction price & Bid & Bidding period\\ \hline
			Successive Outbidding (SO)& Successively outbids himself even though he is the current winner& Bid & Bidding period\\ \hline
			Last Bidding (LB)& Becomes inactive at the last stage to avoid winning& Bid & Last bidding stage\\ \hline	
			Early Bidding (EB)& Tends to bid pretty early in the auction to get users attention & Bid & Early bidding stage\\ \hline
			Winning Ratio (WR)& Competes in many auctions but hardly wins any auctions & Bidder & User history\\ \hline	
			Auction Bids (AB)& Tends to have a much higher number of bids than the average of bids in auction dataset & Auction &Auction history\\ \hline
			Auction Starting Price (ASP)& Offers a small starting price to attract genuine bidders & Auction & Auction history \\  
		\hline							
	\end{tabular}
	\label{SBPatterns}
\end{table}

\section{Production of Shill Bidding Data from Online Auctions}
\subsection{Auction Data Extraction and Preprocessing}
To obtain a reliable SB training dataset, it must be built from actual auction data. Nevertheless, producing high quality auction data is itself a burdensome operation due to the difficulty of collecting data from auction sites on one hand, and the challenging task of preprocessing the raw data on the other hand. The latter consumes a significant time and effort, around  60\% to 80\% of the entire workload \cite{ravisankar2011detection}.  In our previous study \cite{alzahrani2018scraping} , we employed the professional scraper Octopars \footnote{https://www.octoparse.com} to collect a large number of auctions for one of the most popular products on eBay. The extracted dataset contains all the information related to auctions, bids and bidders.  We crawled completed auctions of the iPhone 7 for three months (March to June 2017). We chose iPhone 7 because it may have attracted malicious moneymakers due to the following facts: 

\begin{itemize} 
\item Its auctions attracted a high number of bidders and bids.
\item It has a good price range with the average of \$610.17 (US currency).  Indeed, there is a direct relationship between SB fraud and the auction price \cite{Dong2009}.
\item The bidding duration varies between 1 (20.57\%), 3 (23.2\%), 5 (16.23\%), 7 (38.3\%) and 10 (1.7\%) days. In long duration, a dishonest bidder may easily mimic usual bidding behaviour \cite{Dong2009}. However, as claimed in \cite{chang2014}, fraudulent sellers may receive positive rating in short duration. Thus, we considered both durations. 
\end{itemize}

Table \ref{Statistics1} presents the statistics after preprocessing the scraped auction data.  This operation was very time consuming as it required several manual operations \cite{alzahrani2018scraping}: 1) removing redundant and inconsistent records, and also records with missing bidder IDs; 2) merging several attributes into a single one; 3) converting the format of several attributes into a proper one; 4) assigning auction IDs.  For instance, Date and Time attributes in each auction are converted into seconds; e.g. 1 day and 10 day durations are converted into 86,400 and 864,000 seconds respectively. 

\begin{table}[H]
		\centering
			\caption{Preprocessed Auctions of iPhone 7}
			\begin{tabular}{|p{1.5in}|p{.5in}|}
				\hline
				                  	No. of Auctions & 807 \\ \hline				
              		No.  of Records & 15145 \\ \hline					
                	No.  of Bidder IDs & 1054  \\ \hline
				 	No.  of Seller IDs & 647 \\ \hline
				 	Avg. Winning Price & \$ 578.64  \\ \hline	
				 	Avg. Bidding Duration & 7  \\ \hline
					No. of Attributes & 12  \\	
               		\hline							
			\end{tabular}
			\label{Statistics1}
		\end{table}
	
The metrics of SB patterns are presented in \cite{Sadaoui2017, alzahrani2018scraping}. Each metric is scaled to the range of [0, 1]; a high value indicates suspicious bidding behaviour.
We computed each metric against each bidder in each of the 807 auctions  \cite{alzahrani2018scraping}.  As a result, we obtained a SB training dataset with a total of 6321 instances. Each instance denotes the conduct of a bidder in a certain auction. An instance is a vector of 10 elements: Auction ID, Bidder ID and the eight SB features.

\section{Hierarchical Clustering of Shill Bidding Data}
Since SB data are not labeled, data clustering, an unsupervised learning method, can be utilized to facilitate the labeling task.
Clustering is the process of isolating instances into K groups based on their similarities. The clustering techniques fall into one of the following categories:  1) Partitioning-based, such as  K-medoids and K-means; 2) Hierarchical-based, such as BIRCH, GRIDCLUST and CURE; 3) Density-based, such as DBSCAN and DBCLASD; 4) Grid-based, such as STING and CLIQUE. In our work, we select agglomerative (bottom-top) Hierarchical Clustering (HC) where instances are arranged in the form of a  tree structure using a proximity matrix.
 
\subsection{CURE Overview}
Among the hierarchical clustering methods, we choose CURE  because it is very performant in handling large-scale multi-dimensional datasets, determines non-spherical shapes of clusters, and efficiently eliminates outliers \cite{guha1998cure}.  Random sampling and partitioning techniques are utilized to handle the large-scale problem. Each selected instance is first considered as an individual cluster, and the clusters with the closest distance/similarity are integrated into a final cluster \cite{guha1998cure}. Two novel strategies have been introduced in CURE:
\begin{itemize}
\item { \bf Representative Points (RPs)}, which are selected data points that define the cluster boundary. Instead of using a centroid, clusters are identified by a fixed number of RPs that are well scattered. Clusters with the closest RPs are merged into one cluster. The multiplicity of RPs allow CURE to obtain arbitrary clustering shapes. 
\item {\bf Constant shrinking factor ($\alpha$)}, which is utilized to shrink the distance of RPs towards the centroid of the cluster. This factor reduces noise and outliers. 
\end{itemize}

The worst case time complexity of CURE is estimated to $O(N^2\ log\ N)$, which is high when $N$ is large ($N$ is the number of instances) \cite{xu2015comprehensive}. Since SB data clustering is an offline operation, so the running time is not an issue. The only disadvantage of CURE is that the two parameters RP and $\alpha$ have to be set up by users.  To run the experiments, we utilize the Anaconda-Navigator environment for running Python 3, and we integrate GitHub service to incorporate the CURE program developed by Freddy Stein and Zach Levonian \footnote{https://github.com/levoniaz/python-cure-implementation/blob/master/cure.py}. 
%\vspace{-1.5cm}

\subsection{SB Data Preparation}
Since the bidding duration is used as a denominator in EB and LB patterns, the large gap between different durations greatly affects the computation results. The pattern value for 10 days is far smaller than for 1 day.  So, before applying CURE, we first partition the SB dataset into five subsets according to the five durations (1, 3, 5, 7 and 10 days) as presented in Table \ref{Auctions_Attributes}. We compute the Mean and STandard Deviation (STD) of each subset, which will be individually passed into CURE for clustering along with the best number of clusters. Here a cluster contains instances/bidders with similar bidding behaviour. 

%No. of Auctions & 166 & 187 & 131 & 309 & 14	\\ \hline
%No. of Instances & 1289 & 1408 & 1060 & 2427 &	 137\\ \hline
\begin{table}[H]
	\centering
	\caption{SB Dataset Partitioning According to Bidding Duration}
	\begin{tabular}{|p{0.55in}|p{0.3in}|p{0.33in}|p{0.34in}|p{0.33in}|p{0.37in}|}
		\hline
			Partition & 1 Day & 3 Days & 5 Days & 7 Days & 10 Days \\ \hline					
			\multicolumn{6}{|c|}{{\bf Mean of each pattern per partition}} \\\hline
			BT & 0.1434 & 0.1394 & 0.1419 & 0.1455 & 0.1162	\\ \hline
			BR & 0.1287 & 0.1328 & 0.1235 & 0.1273	 & 0.1021	\\ \hline
			SO & 0.0996 & 0.1047 & 0.0872 & 0.1149	 & 0.0620	\\ \hline	
			LB & 0.4624 & 0.4511 & 0.4676 & 0.4678 & 0.4746 \\ \hline	
			EB & 0.4314 & 0.4192 & 0.4318 & 0.4348	 & 0.4575	\\ \hline
			WR & 0.3812	& 0.3718 & 0.3810 & 0.3533 & 0.3496	\\ \hline
			AB & 0.2120 & 0.1936 & 0.2403 & 0.2567 & 0.2926	\\ \hline
			ASP & 0.5007 & 0.4301	& 0.4478 & 0.4801 & 0.7123	\\ \hline
			Avg. Means & 0.2949 & 0.2802	& 0.2901 & 0.2975 & 0.3208	\\    
		\hline			
			\multicolumn{6}{|c|}{{\bf STD of each pattern per partition}} \\\hline
			BT & 0.1973 & 0.1884 & 0.1984 & 0.2019 & 0.1811	\\ \hline
			BR & 0.1246 & 0.1330 & 0.1243 & 0.1377 & 0.1165	\\ \hline
			SO & 0.2764 & 0.2811 & 0.2583 & 0.2917 & 0.2215	\\ \hline	
			LB & 0.3773 & 0.3753 & 0.3917 & 0.3783 & 0.3931 \\ \hline	
			EB & 0.3775 & 0.3742 & 0.3921 & 0.3802 & 0.3968	\\ \hline
			WR & 0.4356	& 0.4373 & 0.4402 & 0.4345 & 0.4398	\\ \hline
			AB & 0.2323 & 0.2426 & 0.2646 & 0.2658 & 0.2575	\\ \hline
			ASP & 0.4931 & 0.4831 & 0.4863 & 0.4908 & 0.4510	\\ \hline
			Avg. STDs & 0.3142 & 0.3143 & 0.3194 & 0.3226 & 0.3071	\\   
			\hline				
	\end{tabular}
	\label{Auctions_Attributes}
\end{table}

\subsection{Optimal Number of Clusters}
\begin{figure}[h]
\centering
\includegraphics[scale=.4]{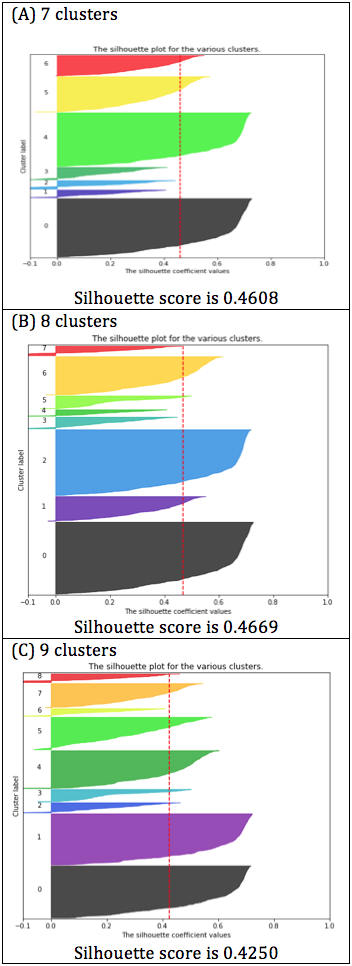} 
\caption{Optimal number of clusters for 7 day bidding duration. Silhouette score is examined 19 times. We show the top three Silhouette scores.}
\label{Number-of-Clusters}
\end{figure}

It is always difficult to decide about the optimal number of clusters. Besides, normal and shill bidder behaviour is somehow  similar. Thus, determining the best number of clusters is an essential step to achieve a better interpretation for classifying similar SB instances \cite{yu2014automatic}. There are several methods to address this problem, such as Elbow, Dendrogram, the Rule of Thumb and Silhouette. In our study, we employ the Silhouette method where each group is represented as a silhouette based on the separation between instances and the cluster's tightness \cite{rousseeuw1987silhouettes}. The construction of a silhouette requires the clustering technique to generate the partitions and also to collect all approximates between instances \cite{rousseeuw1987silhouettes}.  K-mean clustering algorithm has been successfully utilized for this task due to its simplicity and effectiveness \cite{yu2014automatic}. Therefore, we apply K-mean to estimate the number of clusters for each of the five SB subsets. Next, we examine the silhouette scores for 19 clusters, and choose the best number based on the best silhouette score.  In Figure \ref{Number-of-Clusters}, we give an example for the 7 day bidding duration.  Instances are partitioned into eight clusters since the highest silhouette score (0.4669) is obtained on that number. In Table \ref{bestNumberofC}, we expose the best number of clusters for each of the five SB subsets.

\begin{table}[h]
	\centering
	\caption{The best number of clusters for each SB subset}
	\begin{tabular}{|p{0.5in}|p{1in}|}
	\hline
         {\bf Subset}	& {\bf Number of Clusters}  \\ \hline
         1 Day & 7   \\ \hline					
         3 Days & 7  \\ \hline
         5 Days & 5  \\ \hline
         7 Days & 8  \\ \hline
		 10 Days & 2	 \\ \hline
		 Total & 29  \\ 
         \hline							
	\end{tabular}
\label{bestNumberofC}
\end{table}

\subsection{Cluster Generation}
\label{decisionLine}
CURE has three parameters that need to be setup: representative points, shrinking factor and optimal number of clusters. Based on the results of silhouette, we obtain the optimal number of  clusters for each of the five SB subsets. RPs and $\alpha$ are defined by selecting the setting that provides the best instance distribution among the specified clusters.  Thus, CURE is applied with different RPs and $\alpha$ parameters starting from the default values on the eight clusters (see Table \ref{7DaysClustering}).

\begin{table*}[h]
	\centering
	\caption{CURE Clustering for 7 day Bidding Duration}
	\begin{tabular}{|p{0.3in}|p{0.3in}|p{0.3in}|p{0.3in}|p{0.3in}|p{0.3in}|p{0.3in}|p{0.3in}|p{0.3in}|p{0.3in}|}
	\hline
		{\bf RP} & {\bf $\alpha$} & {\bf C\#1} & {\bf C\#2} & {\bf C\#3} & {\bf C\#4} & {\bf C\#5} & {\bf C\#6} & {\bf C\#7} & {\bf C\#8} \\ 
		\hline
         5 & 0.1 & 136 & 1438 & 1 & 2 & 657 & 190 & 2 & 1 \\ \hline
         5 & 0.05 & 657 & 328 & 2 & 1408 & 1 & 28 & 2 & 1 \\ \hline					
         5 & 0.01 & 1438 & 640 & 1 & 1 & 17 & 1 & 1 & 328 \\ \hline
         5 & 0.001 & 21 & 166 & 2 & 1 & 2 & 25 & 2209 & 1 \\ \hline
         10 & 0.1 & 2 & 1 & 657 & 1410 & 22 & 8 & 137 & 190 \\ \hline
         10 & 0.05 & 2 & 133 & 2 & 1 & 2 & 31 & 2066 & 190 \\ \hline
         10 & 0.01 & 1 & 1 & 1 & 1 & 135 & 31 & 2067 & 190  \\ \hline	
         10 & 0.001 & 189 & 654 & 1410 & 3 & 137 & 31 & 1 & 2 \\ 
         \hline							
	\end{tabular}
\label{7DaysClustering}
\end{table*}

\begin{table}[H]
	\centering
	\caption{SB Instances and CURE Classification}
	\begin{tabular}{|p{0.6in}|p{0.3in}|p{0.3in}|p{0.3in}|p{0.3in}|p{0.3in}|}
		\hline
		   	AuctionID & 900 & 900& 2370& 432 & 1370 \\ \hline
		   	BidderID & z***z & k***a & g***r & 0***0 & o***- \\ \hline
			BT & 0.75 & 0.4705 & 0.8333 & 0.5 & 0.04615 \\ \hline
			BR & 0.3461 & 0.3076 & 0.2 & 0.3333 & 0.0857	\\ \hline
			SO & 1 & 0	& 1 & 0 & 0.5 \\ \hline	
			LB & 0.5667 & 0.1909 & 0.0350 & 0.2199 & 0.2966 \\ \hline	
			EB & 0.5409 & 0.1909 & 0.0239 & 0.0043 & 0.2060 \\ \hline
			WR & 0.75 & 	0.4 & 1 & 0.5 & 0 \\ \hline
			AB & 0 & 0	& 0.3333 & 0 & 0.0526\\ \hline
			ASP & 0 & 0 & 0.9935 & 0 & 0\\ \hline \hline
			\rowcolor{lightgray}
			{\bf Label} & {\bf 1} & {\bf 0}	& {\bf 1} & {\bf 0} & {\bf 0} \\
		\hline							
	\end{tabular}
	\label{Patterns_Example}
\end{table}

\section{Labeling Shill Bidding Data}

 \begin{algorithm}
   \caption{: Labeling Bidders in a Cluster}
   \label{Labeling Shill Bidding Data}
    \begin{algorithmic}[1]
      \Require Mean and STD of the corresponding subset
        		\State Compute $MeanOfCluster$
        		   \If {($MeanOfCluster \geq$ ($Mean + \frac{STD}{2}$))}
      	    		\For{x=1 to $NumberOBiddersCluster$}
      	    			\State $LabelOfBidder_{x}$ = 1 (Suspicious)
      	    		\EndFor
        		    \Else
      	     		\For{x=1 to $NumberOfBiddersCluster$}
      	    			\State $LabelOfBidder_{x}$ = 0 (Normal)
      	    	        \EndFor
        		\EndIf 
\end{algorithmic}
\end{algorithm}

\begin{figure}[H]
\centering
\includegraphics[scale=.3]{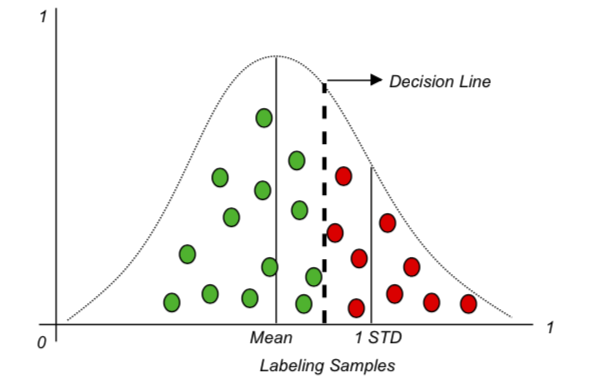} 
\caption{Labeling decision line of a cluster of a subset}
\label{Decision-Line}
\end{figure}

In algorithm \ref{Labeling Shill Bidding Data}, we show the steps to label the bidders of a cluster. A cluster belongs to a certain subset. We consider the value of $Mean + \frac{1}{2} STD$ of the subset since it produces the best decision line that separates between normal and suspicious instances as depicted in Figure \ref{Decision-Line}. The decision line is defined by the average of the means and STDs of the values of the SB patterns in that subset. So, if the mean of the cluster is greater than the decision line, then instances are labeled as suspicious (1) in that cluster; otherwise, they are labeled normal (0).  We also give an example in Table \ref{Patterns_Example} where the bidders belong to the 7 Day subset. As we can observe in this table, the shill bidding instances are similar to the normal bidding instances, and our labeling approach has successfully classified each instance.

Table \ref{statistics3} provides all the best results of the clustering and labeling of our SB dataset. There are 5646 instances categorized as normal and 675 instances as suspicious. The total number of produced clusters is 29.

\begin{table}[h]
	\centering
	\caption{Final Results of Clustering and Labeling}
	\begin{tabular}{|p{.42in}|p{0.23in}|p{0.23in}|p{0.23in}|p{0.23in}|p{0.23in}|p{0.2in}|}
	\hline
         Partitions	& 1 Day & 3 Days & 5 Days & 7 Days & 10 Days & Total \\ \hline
         Auctions & 166 & 187 & 131 & 309 & 14 & 807  \\ \hline					
         Instances & 1289 & 1408 & 1060 & 2427 & 137 & 6321 \\ \hline
         Clusters & 7 & 7 & 5 & 8 & 2 & 29 \\ \hline
         RP & 5 & 5 & 5 & 10 & 5 & NA \\ \hline
	$\alpha$ & 0.05	 & 0.01 & 0.05 & 0.001 & 0.1 & NA \\ \hline
	Normal  & 1135 & 1303 & 975 & 2098 & 135 & 5646 \\ \hline	
	Suspicious & 154 & 105 & 85 & 329 & 2 & 675 \\ 
         \hline							
	\end{tabular}
\label{statistics3}
\end{table}

\section{Conclusion and Future Work}
There are limited SB classification studies due to the difficulty of identifying useful SB strategies in one hand, and the unavailability of labeled training SB data on the other hand. Producing and labeling SB training data are both critical tasks for implementing SB classifiers. Our aim is to effectively label SB data based on hierarchical clustering. CURE has shown a remarkable capability for partitioning the behaviour of bidders. After data clustering, we automatically label each cluster with our own approach.  Our future work mainly concentrates on the following work:

\begin{itemize}
\item The generated SB dataset is highly imbalanced, which will negatively impact the performance of classifiers as shown in \cite{ganguly2017classification}. The decision boundary of the fraud classifiers will be biased towards the normal class, which means suspicious bidders will poorly be detected. Handling the class imbalance problem is a continuous area of study \cite{Zhang2015}. In our research, we will investigate this problem and test different types of techniques, such as over-sampling, under-sampling and cost-sensitive learning, to verify the most suitable technique for our SB dataset.

\item Ensemble learning has produced reliable classification performance for numerous practical applications. The objectives defined by ensemble learning are lowering the model's error ratio, avoiding the overfitting problem, and reducing the bias and variance errors. The most common ensemble methods are Boosting and Bootstrap Aggregation (Bagging). Thus, we will employ the ensemble learning to implement a robust SB detection model, and investigate the most fitting ensemble methods for the SB dataset.
\end{itemize}

\section*{Acknowledgments}

The first author would like to thank Umm Al-Qura University and the Saudi Arabian Cultural Bureau in Canada for the generous financial support. 

\bibliographystyle{IEEEtran}
\bibliography{Paper2References}
\end{document}